%% file: ms.tex
\documentclass{article}
\usepackage{arxiv}
\usepackage[utf8]{inputenc} 
\usepackage[T1]{fontenc}    
\usepackage[bookmarks=false]{hyperref}
\usepackage{url}

\usepackage{graphicx}
\usepackage{amsmath}

\usepackage{subcaption}
\usepackage{float}
\usepackage{subfiles}

\usepackage{multicol}
\usepackage{xcolor}

\author{Peter Hviid Christiansen\\
EIVA\\
{\tt\small pch@eiva.com}
\And
Mikkel Fly Kragh\\
Aarhus University\\
{\tt\small mkha@eng.au.dk}
\And
Yury Brodskiy\\
EIVA\\
{\tt\small ybr@eiva.com}
\And
Henrik Karstoft\\
Aarhus University\\
{\tt\small hka@eng.au.dk}
}

\begin{document}

\title{UnsuperPoint: End-to-end Unsupervised Interest Point Detector and Descriptor}

\maketitle
\subfile{sections/abstract}


\begin{multicols}{2}
\subfile{sections/s01_introduction}

\subfile{sections/s02_network_architecture}
\subfile{sections/s03_self_supervised_framework}
\subfile{sections/s04_loss_functions}
\subfile{sections/s05_experiments}
\subfile{sections/s07_discussion_conclusion_acknowledge}

{\small
\bibliographystyle{unsrt}
\bibliography{ms}
}
\end{multicols}
\end{document}

%% file: sections/abstract.tex
\begin{abstract}
It is hard to create consistent ground truth data for interest points in natural images, since interest points are hard to define clearly and consistently for a human annotator. This makes interest point detectors non-trivial to build. 
In this work, we introduce an unsupervised deep learning-based interest point detector and descriptor. Using a self-supervised approach, we utilize a siamese network and a novel loss function that enables interest point scores and positions to be learned automatically. 
The resulting interest point detector and descriptor is UnsuperPoint. 
We use regression of point positions to 1) make UnsuperPoint end-to-end trainable and 2) to incorporate non-maximum suppression in the model.  Unlike most trainable detectors, it requires no generation of pseudo ground truth points, no structure-from-motion-generated representations and the model is learned from only one round of training. Furthermore, we introduce a novel loss function to regularize network predictions to be uniformly distributed.
UnsuperPoint runs in real-time with 323 frames per second (fps) at a resolution of $224\times320$ and 90 fps at $480\times640$. 
It is comparable or better than state-of-the-art performance when measured for speed, repeatability, localization, matching score and homography estimation on the HPatch dataset. 
\end{abstract}

%% file: sections/s01_introduction.tex
\section{Introduction}
Deep learning \cite{LeCun2015-oh} has since 2012 \cite{Krizhevsky2012-xg} improved a broad range of computer vision tasks. Especially supervised image classification and recognition have reached (super-)human-level performance \cite{He2015-pm, Esteva2017-zy, Chung2016-zw, Rajpurkar2017-ue}.
In particular, deep learning-based methods have improved and influenced traditional tasks in geometric computer vision \cite{Hartley2003-gk} such as pose estimation \cite{Kendall2015-ql, Kendall2017-iv}, homography estimation \cite{DeTone2016-qe}, stereo matching \cite{Luo2016-vm} and visual odometry \cite{Wang2017-rh}. 
Furthermore, deep learning-methods have powered new applications that previously did not exist such as depth from monocular camera \cite{Godard2017-hx}  and pose estimation, where position and orientation are estimated directly using regression \cite{Kendall2015-ql}. 
Nevertheless, traditional interest point detectors \cite{Tareen2018-js}  (SIFT \cite{Lowe2004-qf}, SURF \cite{Bay2008-nf}, ORB \cite{Rublee2011-ns}, AKAZE \cite{Alcantarilla2013-au}, BRISK \cite{Leutenegger2011-ar}) are still commonly used in practical applications, where the concept of points and descriptors remains a powerful representation - in particular because interest point correspondences for a set of images can be established by both ensuring that points match by their descriptors and that matching point positions also satisfy multi-view geometric constraints. Point correspondences are also key in bundle adjustment \cite{Triggs1999-lt}  used in Structure-from-Motion (SfM), Photogrammetry, Visual Simultaneous Localization and Mapping (VSLAM) and Augmented Reality (AR). Bundle adjustment enables large-scale applications (long sequence recordings in large scenes) \cite{Mur-Artal2015-mm, Mur-Artal2017-om}, correction of maps based on loop closure \cite{Mur-Artal2014-cd}, and fusion with odometry sensors such as GPS \cite{Lhuillier2012-le} or as in Visual-Inertial SLAM \cite{Leutenegger2013-kj, Qin2018-wd}, where an IMU is used to reduce drift and improve localization. 

In recent years, deep learning-based interest point detectors and descriptors have gained popularity. However, most research only address the descriptor, for discriminating local image patches \cite{Han2015-vc, Brown2011-ml, Simo-Serra2015-lv, Zagoruyko2015-gz, Balntas2016-nk, Tian2017-pw} as defined in e.g. the Brown \cite{Brown2011-ml} dataset. These methods do not address detection of points, and they rely on traditional interest point detectors. 

The challenge for learning point detectors is that valid ground truth data for interest points in natural images are hard to create. The definition of an interest point is not clearly defined and consistent labels by a human annotator are hard to acquire. The lack of ground truth data thus makes point detectors non-trivial to train.

\subsection{Related work}
In TILDE \cite{Verdie2014-zw}, pseudo ground truth interest points are obtained by selecting points from a Difference-of-Gaussians blob detector \cite{Lowe2004-qf}  that are repeatable across an image sequence. Each sequence is captured from the same viewpoint at different times of day and at different seasons. The drawback is that the detector is trained on static viewpoint images. That is, it is not trained explicitly for rotation and scale invariance. 

Quad-network \cite{Savinov2017-ew} uses unsupervised learning for training a shallow neural network for interest point detection. The model is trained to learn a ranking of points that are preserved under image transformations. This enables the model to learn what defines a good interest point. However, the model runs only on patches and does not provide descriptors for each patch. 

LIFT \cite{Yi2016-on} is able to predict both points and descriptors using three modules; a detector to create a score map of good interest points, an orientation estimator to predict the orientation of a patch and a descriptor module. The score map is used to crop patches of good interest points. A Spatial Transformer Network (STN) \cite{Jaderberg2015-xw} rotates patches by the estimated orientation before a descriptor is created for each patch. LIFT is end-to-end differentiable on patches. However, the model does not train on whole images and does not converge when trained from scratch. It is trained in multiple steps and requires an SfM pipeline to guide the training initially. Furthermore, each module in the LIFT framework does not share computations, making it too slow for real-time applications.

LF-Net \cite{Ono2018-zq} uses (similar to LIFT) a module for selecting good image patches and transforms them using an STN before passing patches through a descriptor module. The training framework is end-to-end differentiable and learns both positions, scale, rotation and descriptors. Unlike LIFT, position, rotation and scale are estimated by a single module. LF-Net is able to train on full images from scratch, it is fast and the model has demonstrated state-of-the-art performance for SfM image matching. However, the framework requires the output of an SfM pipeline during training and it does not share computations between the detector and the descriptor. The use of patches also restricts the area from which the network is able to learn descriptors. 

SuperPoint \cite{DeTone2018-yw} is also able to predict both points and descriptors. However, in SuperPoint the detector and descriptor share most computations, making it fast. SuperPoint is trained using online-generated synthetic images of simple geometrical shapes. Pseudo ground truth points are defined as corners, junctions, blobs and line segments of the synthesized data. However, to generalize for “real” images, two rounds of homography adaptation are used. The model is initially trained on the synthesized data, followed by training on real images to generate pseudo ground truth interest points by aggregating predictions of 100 different homography transformations per image. A new model is trained and the homography adaptation step is repeated to improve pseudo ground truth points even further. Finally, a new model is trained to predict both points and descriptors using a siamese network. However, the initial two/three rounds of training are cumbersome and a good interest point is initially only defined by the authors as junctions of simple geometrical shapes in the synthetic data. 

In this work, we present UnsuperPoint - a fast deep learning-based interest point detector and descriptor inspired by SuperPoint \cite{DeTone2018-yw}. Similar to SuperPoint, the model shares most computations for the detector and the descriptor and utilizes a siamese network to train descriptors. However, in UnsuperPoint, we use regression for predicting positions and introduce a novel interest point detector loss function to also train detection of points in a self-supervised manner. Unlike most trainable detectors, it requires only one round of training, no generation of pseudo ground truth points and no SfM-generated representations. Finally, we also introduce a novel loss function to easily regularize network predictions to be uniformly distributed.

%% file: sections/s02_network_architecture.tex
\section{Network architecture}
\begin{figure*}
  \centering
    \includegraphics[width=1.00\textwidth]{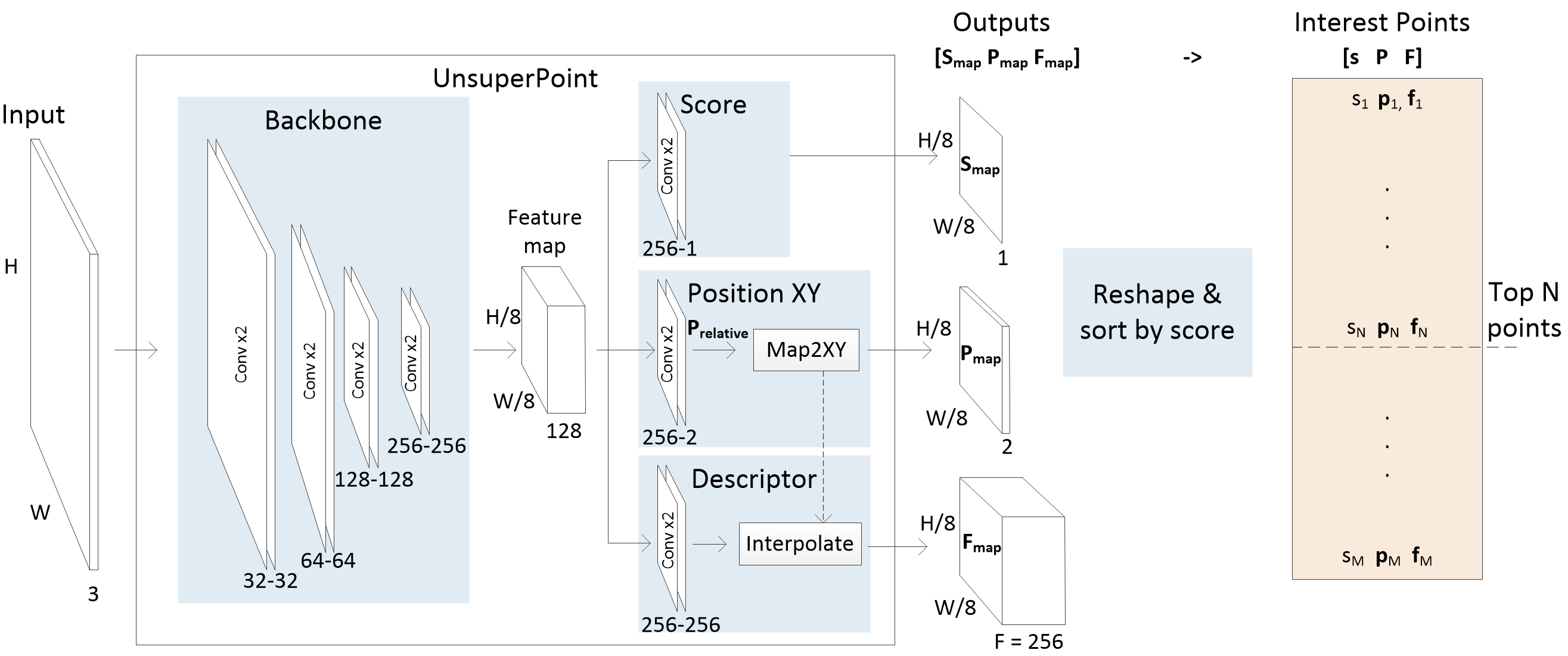}
  \caption{UnsuperPoint takes an input image and outputs an interest point vector. The score, position and descriptor share most computations by a shared backbone. Each interest point $m$ is described by a score $s_{m}$, a position $\mathbf{p}_{m}$ and a descriptor $\mathbf{f}_{m}$. }
  \label{fig:architecture}
\end{figure*}

UnsuperPoint has a multi-task network architecture with a shared backbone followed by multiple task-specific submodules as shown in Figure \ref{fig:architecture}. 

The backbone takes a color image as input and provides a downsampled feature map that is further processed by task-specific submodules in the same way as done in Superpoint \cite{DeTone2018-yw}. The submodules process the backbone output with additional convolutional layers. 
The convolutional structure of backbone and subtasks enables the model to process any input image size. The subtasks are designed to produce an aligned output where each entry represents a point with a position, score and descriptor.

The combined output of the network resembles the output of traditional point detectors by providing a position, score and descriptor for each interest point. Thus, the network can be used as a drop-in replacement for traditional interest point-based systems such as SfM, AR and VSLAM. 

\subsection{Network overview and notation}
Each point position is expressed by its relative position $\mathbf{P}_\text{relative}$ and is easily transformed to image pixel coordinates $\mathbf{P}_\text{map}$. 
The score $\mathbf{S}_\text{map}$ is the fitness of each point and used for sampling the best $N$ points. The descriptor map $\mathbf{F}_\text{map}$ has an embedding of $F$ channels for each entry to uniquely match corresponding points from different images. 
$\mathbf{S}_{map}$, $\mathbf{P}_\text{map}$ and $\mathbf{F}_\text{map}$ are reshaped and sorted by highest score into respectively a vector $\mathbf{s}$ with $M$ elements, an $M \times 2$ matrix $\mathbf{P}$ and an $M \times F$ matrix $\mathbf{F}$, where $M = \frac{H}{8} \cdot \frac{W}{8}$ represents all predicted points. The top $N$ interest points are simply the top $N$ rows of the reshaped output. All convolutional layers have a stride of 1 and a kernel size of 3. Apart from the final layer in each subtask, all convolutional layer are followed by batch normalization \cite{IoffeS14} and a leaky ReLU \cite{He2015-pm} activation function. 

\subsection{Backbone module}
The backbone takes an input image and generates an intermediate feature map representation to be processed by each subtask. The backbone is fully convolutional with four pairs of convolutional layers. The four pairs are separated by three max-pooling layers with a stride and a kernel size of two. After each pooling layer, the number of channels for subsequent convolutional layers are doubled. The number of channels in the eight convolutional layers are 32-32-64-64-128-128-256-256. Effectively, each pooling layer downsamples the feature map height and width by a factor of two, while the whole backbone downsamples by a factor of eight. An entry in the final output corresponds to an $8\times8$ area in the input image. Thus, for an input image of e.g. $480\times640$, the network will return $\left(480/8\right)\cdot\left(640/8\right) = 4800$ entries. Each entry is processed in a fully convolutional way for each subtask to output a descriptor, score and position - effectively creating 4800 interest points. 

\subsection{Score module}
The score module regresses a score for each entry in the final feature map. The score module contains two convolutional layers with 256 and 1 channels, respectively. The final layer is followed by a sigmoid activation to bound score predictions in the interval $\left[0, 1\right]$. The scores are important for selecting the top $N$ points in an image. 

\subsection{Position module}
\begin{figure*} 
  \centering
  \includegraphics[width=0.5\textwidth]{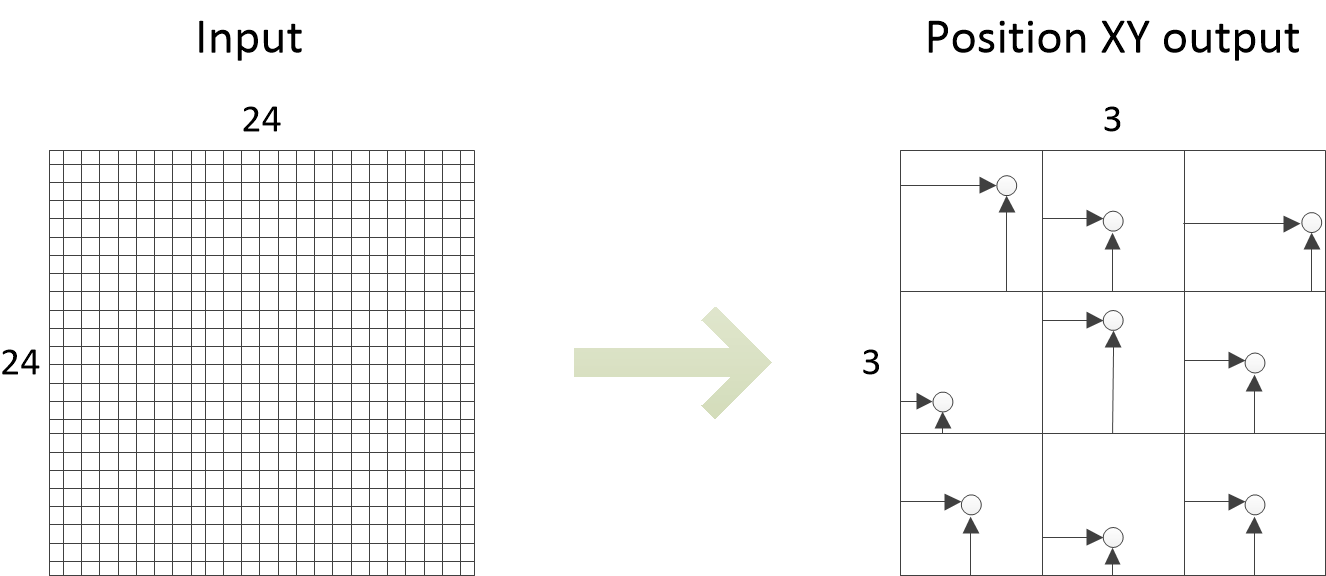}
  \caption{The network predicts a single interest point position for each $8\times 8$ area. For a $24\times 24$ sized input image, the network predicts the position of $3 \cdot 3 =9$ interest points.}
  \label{fig:regression}
\end{figure*}

The position module predicts a relative image coordinate for each output entry and maps this to an image pixel coordinate. The position module contains two convolutional layers with 256 and 2 channels, respectively. The final layer is followed by a sigmoid activation to bound position predictions in the interval $\left[0, 1\right]$. For a network with 3 poolings layers (a subsampling factor of 8), a relative position is predicted for each $8\times8$ region in the input image. This is demonstrated in Figure \ref{fig:regression} for a small input image of size $24\times24$. 

The mapping from relative image coordinates $\mathbf{P}_{\text{relative}}$ to image pixel coordinates $\mathbf{P}_{\text{map}}$ is calculated by:  
\begin{flalign} \label{eq:positionxy_map2xy}
  \begin{aligned}
	\mathbf{P}_{\text{map,x}}\left( r, c\right) = & \left( c +  \mathbf{P}_{\text{relative,x}}\left( r, c\right) \right) \cdot f_\text{downsample}\\
	\mathbf{P}_{\text{map,y}}\left( r, c\right) = & \left( r + \mathbf{P}_{\text{relative,y}}\left( r, c \right) \right) \cdot f_\text{downsample}
  \end{aligned}
\end{flalign}

Relative image coordinates $\mathbf{P}_{\text{relative}}$ are added by the column entry index $c$ for the $x$-coordinate, and the row entry index $r$ for the $y$-coordinate. The output is then multiplied by the downsampling factor $f_\text{downsample} = 8$ of the network.

Using regression for point detection is a clear distinction to SuperPoint \cite{DeTone2018-yw} and LF-Net \cite{Ono2018-zq}, where top interest point locations are selected from a heat map of the same size as the input image. The purpose of using regression for estimating position is two-fold. Most importantly, it is differentiable and enables fully unsupervised training. Secondly, by only predicting a single point for each $8\times8$ area, it adds functionality similar to non-maximum suppression (NMS) as part of the network architecture. Intuitively, NMS might come as a disadvantage, as an $8 \times 8$ region may have multiple interest points. However, by removing closely clustered points, interest points will become more homogeneously distributed. This is a desired property and many interest point based systems use NMS to improve robustness, stability and accuracy \cite{Bailo2018-cp}.

\subsection{Descriptor module}
The descriptor module generates a descriptor for each entry. The descriptor module contains two convolutional layers with 256 and $F = 256$ channels, respectively. The final layer has no activation. The descriptor map can be used coarsely or by interpolating descriptors based on interest point positions. In SuperPoint, the interpolation is a post-processing step used under inference. In our implementation, interpolation of descriptors is integrated into the model. The model use all point positions in $\mathbf{P}_\text{map}$ to interpolate all entries in descriptor map $\mathbf{F}_\text{map}$. Regression of point positions makes interpolation of descriptors differentiable, and it is used in training. 

%% file: sections/s03_self_supervised_framework.tex
\section{Self-supervised framework}
\begin{figure*}
  \centering
    \includegraphics[width=1.0\textwidth]{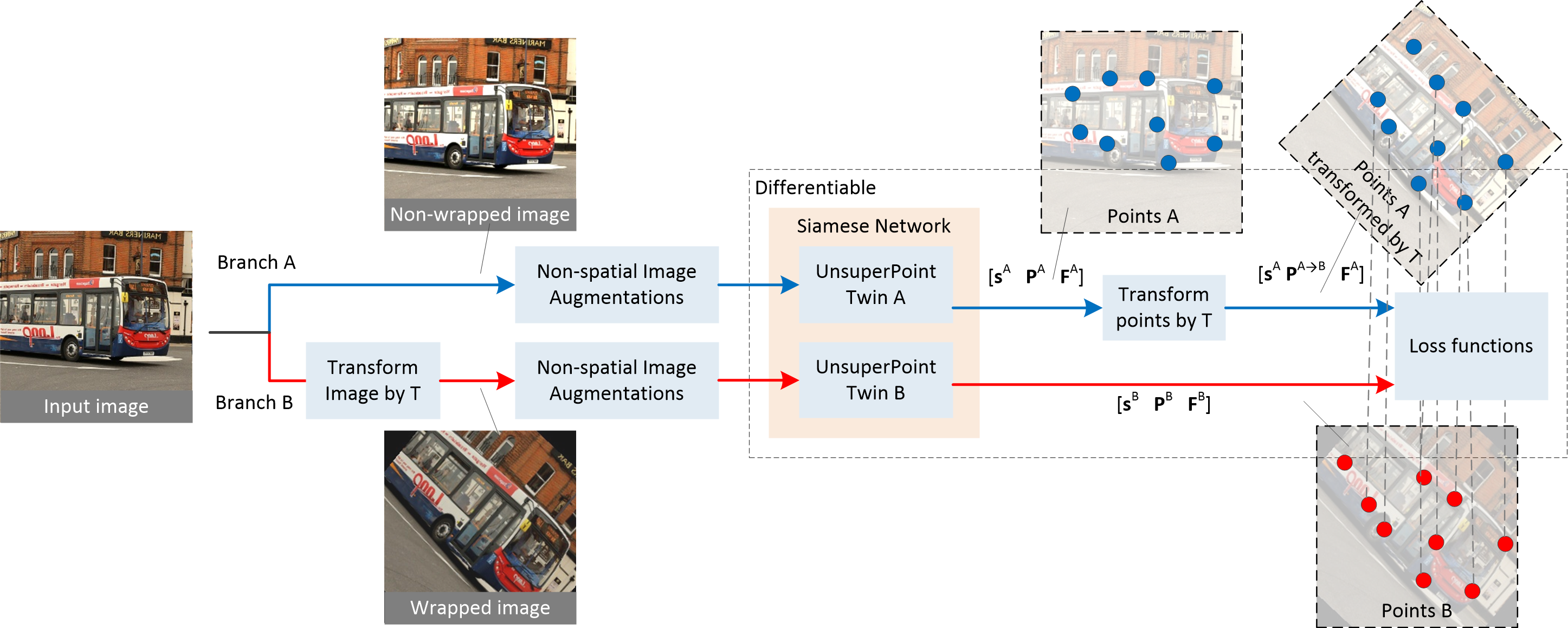}
  \caption{Two permutations of the same image are forwarded through a siamese network. Corresponding points between branch A and B are determined and used for training the model with supervised loss functions.}
  \label{fig:training_framework}
\end{figure*}

UnsuperPoint uses a self-supervised training framework to learn all three tasks simultaneously. The procedure is demonstrated in Figure \ref{fig:training_framework}, where UnsuperPoint is used in a siamese network to predict interest points for two augmentations of the same input image. The two augmentations and their predictions are separated into individual branches. Branch A (blue) is a non-wrapped version of the input image, whereas branch B (red) is a wrapped version of the input image. 

The image in branch B is spatially transformed by a random homography $T$ (rotation, scale, skew and perspective transforms). The image for each branch is then transformed by independent random non-spatial image augmentations such as brightness and noise. UnsuperPoint predicts interest points on the image from each branch. Point positions of branch A are transformed by $T$ to spatially align points from branch A to branch B. We define points from each branch to correspond if they are spatially close after alignment. Finally, point correspondences are used in loss functions to train the model. All components in UnsuperPoint, the transformation of points by $T$ and loss functions are differentiable, thus enabling the model of each branch to be trained end-to-end. 

%% file: sections/s04_loss_functions.tex
\section{Loss functions}\label{sec:loss}
This section presents the loss functions to train score, position and descriptor. 
The total loss $\mathcal{L}_\text{total}$ consists of four loss terms.

\begin{equation}\label{eq:loss_all}
\begin{split}
\mathcal{L}_\text{total} = &\textcolor{white}{+\ } \alpha_{\text{usp}} \mathcal{L}^{\text{usp}} \\ 
	&+\alpha_\text{uni\_xy} \mathcal{L}^\text{uni\_xy} \\ 
	&+\alpha_\text{desc}\mathcal{L}^\text{desc} + \alpha_\text{decorr}\mathcal{L}^\text{decorr}
\end{split}
\end{equation}

Each loss term is weighted by a factor $\alpha$. The first loss term $\mathcal{L}^\text{usp}$ is the UnSupervised Point (USP) loss to learn position and score of interest points. 
The second loss term $\mathcal{L}^\text{uni\_xy}$ 
is a regularization term to encourage a uniform distribution of relative point positions. 
The final two loss terms $\mathcal{L}^\text{desc}$ and $\mathcal{L}^\text{decorr}$ optimize only the descriptor.
$\mathcal{L}^\text{desc}$ is required to learn descriptors, while $\mathcal{L}^\text{decorr}$ is merely a regularization term to reduce overfitting by decorrelating descriptors.

\paragraph{Point-pair correspondences}\label{sec:loss_corresponds}
Each branch $b \in \left\{ A, B \right\}$ outputs three tensors $\mathbf{s}^b$, $\mathbf{P}^b$ and $\mathbf{F}^b$ that contain point scores, point positions and point descriptors. To calculate loss functions in the following sections, we need to establish point correspondences (point-pairs) from the two branches. To do this, an $M^\text{A} \times M^\text{B}$ distance matrix $\mathbf{G}$ is determined by computing the pairwise distances between all $M^\text{A}$ transformed points from branch A and all $M^\text{B}$ points from branch B.

\begin{equation}\label{eq:distance_matrix}
\mathbf{G} = \left[ g_{ij}\right] _{M^\text{A} \times M^\text{B}} = \left[ \left\| \mathbf{p}^{\text{A}\rightarrow \text{B}}_{i} - \mathbf{p}^\text{B}_{j} \right\| _{2}\right] _{M^\text{A} \times M^\text{B}}
\end{equation}

Each entry $g_{ij}$ in $\mathbf{G}$ is the Euclidean distance between a transformed point $\mathbf{p}^{\text{A}\rightarrow \text{B}}_{i} = T\mathbf{p}^\text{A}_{i}$ with index $i$ in branch A and a point $\mathbf{p}^\text{B}_{j}$ with index $j$ in branch B. A point-pair is the combination of a point $i$ in branch A that corresponds to a point $j$ in branch B. Not all points in branch A are merged into point-pairs, because a point in branch A may not have a nearby neighbor in branch B. We define that points correspond if point $i$ in branch A has point $j$ as its nearest neighbor in branch B, and if the distance $g_{ij}$ between these is less than a minimum distance $\epsilon_\text{correspond}$.

With point correspondences, we can redefine output tensors ($\mathbf{s}^b$, $\mathbf{P}^b$ and $\mathbf{F}^b$) to a new set of tensors defined as \textit{corresponding tensors} (respectively $\mathbf{\hat{s}}^{b}$, $\mathbf{\hat{P}}^{b}$ and $\mathbf{\hat{F}}^b$) with $K$ entries, so that each entry $k$ in the \textit{corresponding tensors} maps to the same point in the input image. 
For each entry $k$ in branch $b$, we define a point-pair score $\hat{s}_{k}^{b}$, the point-pair position $\hat{\mathbf{p}}_{k}^{b}$ and the point-pair descriptor $\hat{\mathbf{f}}_{k}^{b}$. 
Finally, we also define the point-pair correspondence distance $d_{k}$ 
written as
\begin{equation}\label{eq:distance_corresponding}
d_{k} = \|T\mathbf{\hat{p}}^\text{A}_{k}-\mathbf{\hat{p}}^\text{B}_{k}\| =  \|\mathbf{\hat{p}}^{\text{A}\rightarrow \text{B}}_{k}-\mathbf{\hat{p}}^\text{B}_{k}\| 
\end{equation}
Similar to equation \ref{eq:distance_matrix}, it is the distance between points from branch A and B. However, $d_{k}$ is only the distance between a point-pair.  


\subsection{Unsupervised point loss, $\mathcal{L}^\text{usp}$}\label{sec:loss_score_point}
We introduce a novel loss function called UnSupervised Point (USP) loss that uses point-pairs to train a detector using an unsupervised loss function. 
The overall objective of the USP loss is to improve repeatability of the detector, meaning that the same interest points are detected - regardless of the camera viewpoint. In other words, the detector should from multiple camera viewpoints predict image positions that capture the same 3D points in the scene.

The unsupervised point loss $\mathcal{L}^\text{usp}$ is divided into three terms and accumulated over all K corresponding point-pairs: 
\begin{equation}\label{eq:loss_unsupervised_point_loss}
\mathcal{L}^{usp} = \alpha_\text{position} \sum_{k = 1}^K   l_{k}^{\text{position}} + \alpha_\text{score} \sum_{k = 1}^K  l_{k}^{\text{score}} + \sum_{k = 1}^K  l_{k}^{\text{usp}}
\end{equation}
We add weight terms to position $\alpha_\text{position}$ and score $\alpha_\text{score}$.

The objective of the first term $l_{k}^\text{position}$ is to ensure that the predicted positions of a point-pair represent the same point in the input image. We can achieve this by simply minimizing the distance for each point-pair $k$:
\begin{equation}
l_{k}^\text{position} = d_{k}
\end{equation}

\begin{figure*}
    \centering
    \begin{subfigure}[b]{0.45\textwidth}
    \includegraphics[width=\textwidth]{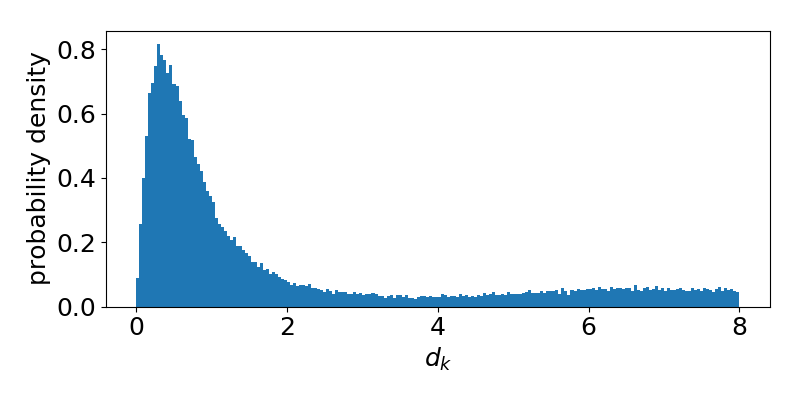}
    \caption{Histogram of point-pairs distances $d_{k}$ }
    \end{subfigure}
    ~
    \begin{subfigure}[b]{0.45\textwidth}
        \includegraphics[width=\textwidth]{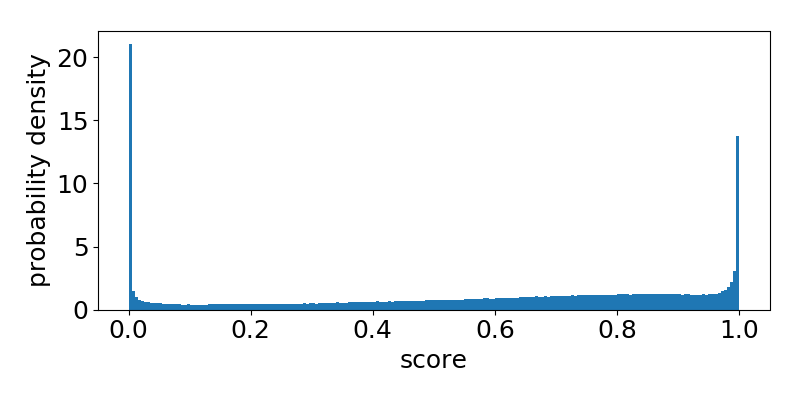}
        \caption{Histogram of scores}
    \end{subfigure}
    \caption{Scores and the distance between point-pairs $d_{k}$.
    \label{fig:score_and_corresponding}
    }
\end{figure*}

Initially, a siamese network will predict random positions. Over time, the siamese network will gradually reduce the distances between point-pairs and thus improve interest point positioning. Figure \ref{fig:score_and_corresponding}a shows an example histogram of point-pair distances for a converged network.

The objective of the second term $l_{k}^\text{score}$ is to ensure that the predicted scores for a point-pair are similar. The second objective is achieved by minimizing the squared distance between score values for each point-pair $k$:
\begin{equation}
l_{k}^\text{score} = \left( \hat{s}_k^\text{A}-\hat{s}_k^\text{B} \right)^2
\end{equation}
In image matching, it is important to have a similar score for points (captured from multiple viewpoints) that represent the same point in the scene. By having similar scores, it is more likely that the N points with the highest score from each image represent the same points in the scene.

The objective of the third term $l_{k}^\text{usp}$ is to ensure that predicted scores actually represent the confidence of interest points. That is, the highest score should be the most repeatable point, and the lowest score should be the least repeatable point. The loss is calculated for each corresponding point-pair $k$:

\begin{equation}\label{eq:loss_unsupervised_point_loss_a}
l_{k}^\text{usp} = \hat{s}_k  \left(d_k-\overline{d}\right)
\end{equation}

Here, $\hat{s}_k$ denotes the joint score of a point-pair and is calculated as 
\begin{equation}
\hat{s}_{k} = \frac{\hat{s}^\text{A}_{k}+\hat{s}^\text{B}_{k}}{2}
\end{equation}
, whereas $\overline{d}$ denotes the average distance between all point-pairs and is calculated as
\begin{equation}
\overline{d} = \frac{1}{K} \sum_{k = 1}^K d_{k}
\end{equation}
The core concept of $l_{k}^{usp}$ is that the network should define a good interest point as a point with a low point-pair correspondence distance $d_k$. 
Oppositely, for a bad interest point, $d_k$ is large, because the network is unable to predict point positions consistently. 

With equation \ref{eq:loss_unsupervised_point_loss_a}, we will have that for $d_{k} < \overline{d}$, the model must learn to set the score high to minimize the loss, and for $d_{k} > \overline{d}$, the model must learn to set the score low to minimize the loss. Figure \ref{fig:score_and_corresponding}b presents how scores are distributed after training. 

%


Effectively, the specified loss function makes the network output increase scores, when the network believes that the same position can be retrieved under the augmentations specified in the framework (spatial and non-spatial augmentations). An advantage of this procedure is that the network is able to learn the characteristics of good interest points based on the provided data and some specified augmentation. The network is free to use both local features (blobs, corners, line segments, textures) and more global features to improve interest point detection.

\subsection{Uniform point predictions, $\mathcal{L}^\text{uni\_xy}$}\label{sec:loss_uniform_xy}
\begin{figure*}
    \centering
    \begin{subfigure}[b]{0.40\textwidth}
        \includegraphics[width=\textwidth]{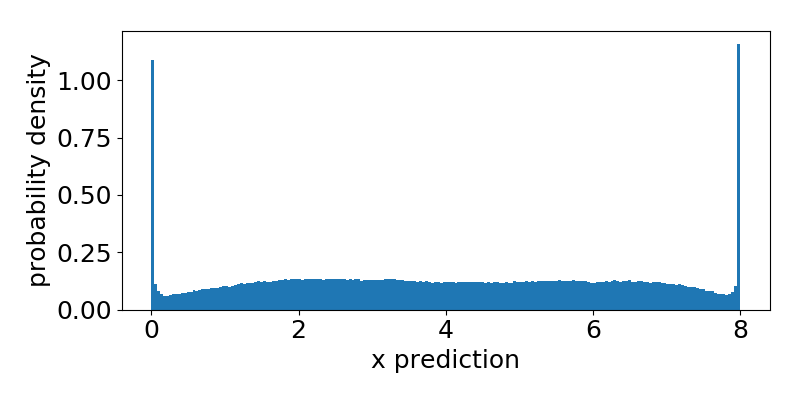}
        \caption{Without uniform regularization}
        \label{fig:x_coordinate_predictions_without_uniform}
    \end{subfigure}
    ~ 
    \begin{subfigure}[b]{0.40\textwidth}
        \includegraphics[width=\textwidth]{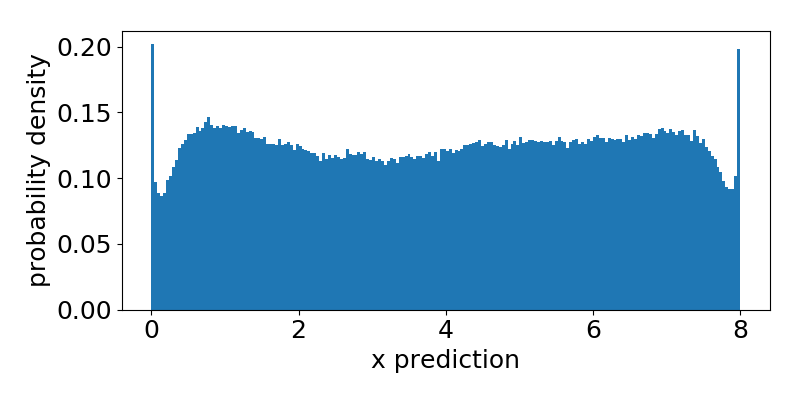}
        \caption{With uniform regularization}
        \label{fig:x_coordinate_predictions_with_uniform}
    \end{subfigure}
    \caption{Histogram of $x$-coordinate position predictions}\label{fig:x_coordinate_predictions}
\end{figure*}

Training a model using only the above-mentioned loss functions introduces some unwanted artefacts for position predictions. Figure \ref{fig:x_coordinate_predictions}a illustrates the artifact in the histogram of predicted x-coordinates.
Recall that the network predicts a relative position in an $8 \times 8$ area in the input image. Optimally, the relative $x$- and $y$-coordinate predictions should be uniformly distributed within this area. However, the histogram in Figure \ref{fig:x_coordinate_predictions}a shows a large number of points near the boundaries (values of 0 and 8). One reason for this is that a model is encouraged, especially for hardly repeatable points, to place the point as closely to points outside its own region in order to minimize $d_{k}$. Instead, it is better to force the model to only predict the best position in its own local region. Thus, we should encourage a uniform distribution of $x$- and $y$-predictions. 

\begin{figure*}
    \centering
    \begin{subfigure}[b]{0.40\textwidth}
        \includegraphics[width=\textwidth]{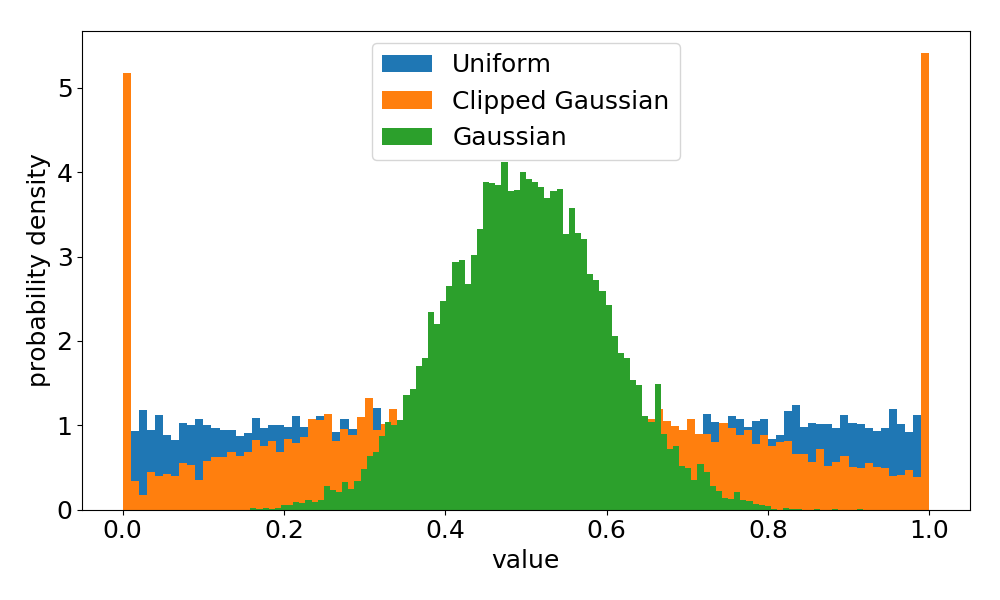}
        \caption{Histogram of distributions}
        \label{fig:uniform_demo_histogram}
    \end{subfigure}
    ~ 
    \begin{subfigure}[b]{0.40\textwidth}
        \includegraphics[width=\textwidth]{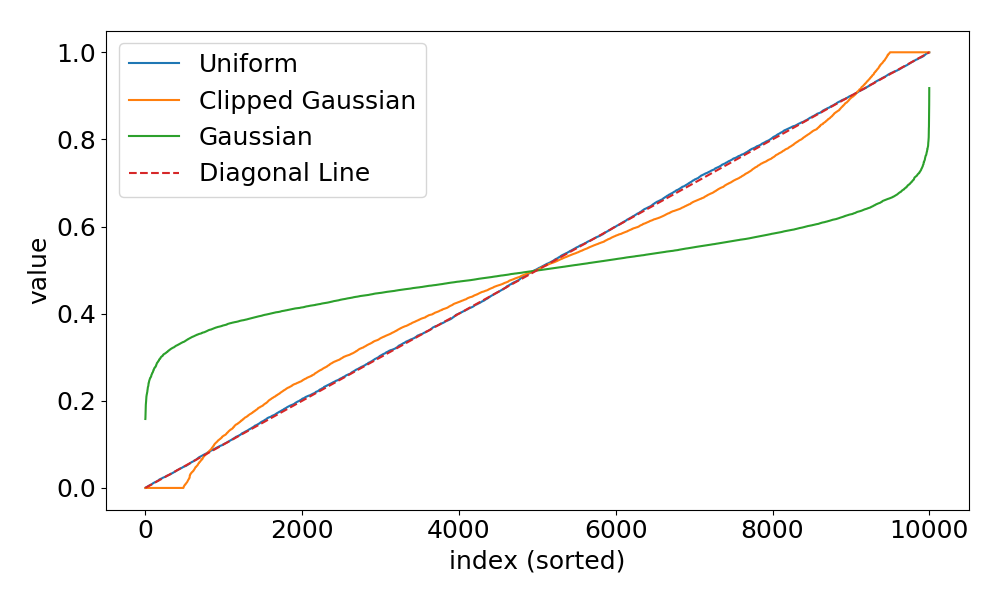}
        \caption{Ascendingly sorted values}
        \label{fig:uniform_demo_sorted}
    \end{subfigure}
    \caption{Uniform, Gaussian and clipped Gaussian distribution centered around 0.5 in the range 0-1.}\label{fig:uniform_demo}
\end{figure*}

We therefore introduce, to our knowledge, a new loss function to encourage a uniform distribution. The core concept is that ascendingly sorted values sampled from a uniform distribution will approximate a straight line going from the lower to the upper bound within the specified range. We demonstrate this by a few simple examples. Figure \ref{fig:uniform_demo}a presents a uniform distribution (blue) and two clipped Gaussian distributions centered around 0.5 with a variance of 0.3 (orange) and 0.10 (green). Figure \ref{fig:uniform_demo}b presents ascendingly sorted samples from the same three distributions. The dashed line is a diagonal line going from lower to upper bound. These examples demonstrate that the distance to a uniform distribution is the distance between sorted values and the diagonal line. 

We define a heuristic measure $D\left(\mathcal{U},  \mathcal{V}\right)$ to calculate the distance between a uniform distribution  $\mathcal{U}\left(a, b\right)$ and some distribution $\mathcal{V}$ in a bound interval $\left[a, b\right]$. $L$ values are sampled from $\mathcal{V}$ to form a vector $\mathbf{v}$. The distance between distributions is defined as
\begin{equation}
D\left(\mathcal{U}\left(a, b\right), \mathcal{V}\right) = \sum_{i=1}^{L}\left( \frac{v_{i}^\text{sorted}-a}{b-a} - \frac{i-1}{L-1}\right)^2
\end{equation}
where $v_{i}^\text{sorted}$ is the ascendingly sorted values of $\mathbf{v}$ such that $v_{i}^\text{sorted} < v_{i+1}^\text{sorted}$ for $i \in \left\{ 1,2, \dots , L-1 \right\}$. 
The first term normalizes sorted values to the interval $\left[0, 1\right]$ and the second term is a proportional line from 0 to 1. For samples in the interval $\left[0, 1\right]$, the loss is 
\begin{equation}
D\left(\mathcal{U}\left(0, 1\right), \mathcal{V}\right) = \sum_{i=1}^{L}\left( v_{i}^\text{sorted} - \frac{i-1}{L-1}\right)^2
\end{equation}
This loss function is a simple and fast procedure to regularizing network predictions to be uniformly distribution. Compared to other methods that measure the difference/distance to a uniform distribution or randomness of a signal (information entropy), this function is differentiable, and it does not require predictions to be discretized. \\

For UnsuperPoint, the loss $\mathcal{L}^\text{uni\_xy}$ is calculated as the distance between a uniform distribution and the distribution of the relative image positions $\mathbf{P}_{\text{relative,x}}$ and $\mathbf{P}_{\text{relative,y}}$, individually. These are denoted $\mathcal{L}^\text{uni\_x}$ and $\mathcal{L}^\text{uni\_y}$, respectively. Unlike the previous section, the loss is calculated independently for each branch, and it does not rely on point correspondences.   
\begin{equation}\label{eq:loss_uniform_xy}
\begin{split}
\mathcal{L}^\text{uni\_xy} &= \alpha_\text{uni\_xy}\left(\mathcal{L}^\text{uni\_x} + \mathcal{L}^\text{uni\_y} \right) 				\\
\mathcal{L}^\text{uni\_x} &= \sum_{i = 1}^M  \left(x_{i}^\text{sorted} - \frac{i-1}{M-1} \right)^2 	\\
\mathcal{L}^\text{uni\_y} &= \sum_{i = 1}^M  \left(y_{i}^\text{sorted} - \frac{i-1}{M-1} \right)^2
\end{split}
\end{equation}
The ascendingly sorted values are respectively $x_{i}^\text{sorted}$ and $y_{i}^\text{sorted}$ for all $M$ points. 
The loss term is weighted by $\alpha_\text{uni\_xy}$.

Figure \ref{fig:x_coordinate_predictions}b presents the distribution of the $x$-coordinates when a model has been trained with a uniform loss. It is clear that the peaks at the boundaries have been reduced significantly. 

\subsection{Descriptor, $\mathcal{L}^\text{desc}$}\label{sec:loss_descriptor}

The descriptor loss is determined using a hinge loss with a positive and a negative margin as described in SuperPoint \cite{DeTone2018-yw}. 

We define an $M^\text{A} \times M^\text{B}$ correspondence matrix $\mathbf{C}$ containing values of either 0 or 1.
Each entry $c_{ij}$ specifies if two points are separated by less than 8 pixels for any pair combination of transformed points in branch A,  $\mathbf{p}^{\text{A}\rightarrow \text{B}}_{i}$ where $i \in \left\{1, 2, \dots, M^\text{A} \right\}$, and points in branch B, $\mathbf{p}^\text{B}_{j}$ where $j \in \left\{1, 2, \dots, M^\text{B} \right\}$. Unlike point-pairs, a single point may correspond to multiple points in the other branch. 

\begin{equation}\label{eq:corresponding_matrix}
c_{ij} = 
\begin{cases}
1 &\text{if}\  g_{ij} \leq 8 \\
0 &\text{otherwise}
\end{cases}
\end{equation}

The hinge loss is calculated using both a positive margin $m_p$ and a negative margin $m_n$ and by accumulating losses for any pair combination of descriptors from branch A and B .

\begin{equation}\label{eq:loss_descriptor}
\begin{split}
\mathcal{L}^\text{desc} &= \sum_{i = 1}^{M^\text{A}} \sum_{j = 1}^{M^\text{B}}  l_{ij}^\text{desc} \\ 
l_{ij}^\text{desc} &=  \lambda_{d}\cdot c_{ij} \cdot \text{max} \left( 0, m_p - {\mathbf{f}_{i}^\text{A}}^{T} \mathbf{f}_{j}^\text{B} \right) \\ & + \left(1- c_{ij} \right) \cdot \text{max} \left( 0, {\mathbf{f}_{i}^\text{A}}^{T} \mathbf{f}_{j}^\text{B} - m_n \right)
\end{split}
\end{equation}

To balance the few corresponding points compared to non-corresponding points, a weight term $\lambda_{d}$ is added. 

\subsection{Decorrelate descriptor, $\mathcal{L}^\text{decorr}$}\label{sec:loss_decorrelate_descriptor}
Feature descriptors are decorrelated to reduce overfitting and improve compactness. Similar to  \cite{Tian2017-pw}, we reduce the correlation between dimensions by minimizing the off-diagonal entries of the descriptor correlation matrix $\mathbf{R}^b = \left[r^{b}_{ij} \right]_{F\times F}$ for each branch $b$. 
\begin{equation}\label{eq:loss_descorrelate_descriptor}
\mathcal{L}^\text{decorr} = \sum_{i \neq j}^{F} \left( r^\text{A}_{ij} \right) +
 \sum_{i \neq j}^{F} \left( r^\text{B}_{ij} \right)
\end{equation}
Each entry $r^{b}_{ij}$ in $\mathbf{R}^b$ is
\begin{equation}\label{eq:loss_correlation}
r^{b}_{ij} = \frac{\left( \mathbf{v}_{j}^{b} -\bar{v}_{j}^{b} \right)^T\left( \mathbf{v}_{i}^{b} -\bar{v}_{i}^{b} \right) }{\sqrt{\left( \mathbf{v}_{j}^{b} -\bar{v}_{j}^{b} \right)^T\left( \mathbf{v}_{i}^{b} -\bar{v}_{i}^{b} \right)}\sqrt{\left( \mathbf{v}_{j}^{b} -\bar{v}_{j}^{b} \right)^T\left( \mathbf{v}_{i}^{b} -\bar{v}_{i}^{b} \right)}}
\end{equation}
where $\mathbf{v}_{i}^{b}$ is a $M^{b} \times 1$ sized vector and is the $i$th column of $\mathbf{F}^{b}$, and $\bar{v}_{i}^{b}$ is the mean of $\mathbf{v}_{i}^{b}$.


%% file: sections/s05_experiments.tex
\section{Experimental details}
The siamese network was trained with PyTorch \cite{Paszke2017-wo}. We used the 118,287 training images from MS COCO \cite{Lin2014-ee}, but without labels. The whole model was trained for 10 epochs with data shuffling, a batch size of 5 (for each branch) and color images of size $240\times320$. Images were normalized by subtracting 0.5 and multiplying with 0.225 for each channel. We used an ADAM optimizer with a default setting as specified in \cite{Kingma2014-gz}. A random homography transformation was constructed for branch A by combining scale, rotation and perspective transformations. The amount of scaling, rotation and perspective transformation was sampled uniformly in restricted intervals. Furthermore, we added standard non-spatial augmentations for each branch such as noise, blur and brightness. The magnitude of each augmentation was uniformly distributed within restricted intervals. The max distance between corresponding points was $\epsilon_\text{correspond} = 4$. We adopted the descriptor loss weights from SuperPoint with a positive margin $m_{p} = 1$, a negative margin $m_{n} = 0.2$ and a balancing factor $\delta_{d} = 250$. 
The search space for estimating the optimal weight for loss terms in equation \ref{eq:loss_all} is large and has  not been investigated. The procedure for selecting appropriate weight terms has been adjusted coarsely for each new loss term added to the total loss. The selected weight terms were $\alpha_\text{usp} = 1$, $\alpha_\text{position} = 1$, $\alpha_\text{score} = 2$, $\alpha_\text{uni\_xy} = 100$, $\alpha_\text{desc} = 0.001$ and $\alpha_\text{decorr} = 0.03$. 

\section{Experiments}
In experiments, we have evaluated configurations of UnsuperPoint to highlight the benefit of interpolating descriptors, adding a loss to uniformly distribute point predictions and adding a loss to decorrelate descriptors. Furthermore, we compare UnsuperPoint to state-of-the-art. 

\subsection{Metrics}
We have used the evaluation metrics from SuperPoint \cite{DeTone2018-yw} by evaluating interest point positions by repeatability rate and localization error, and by evaluating the whole detector (score, position and descriptors) in a homography estimation framework by measuring matching score and the homography accuracy. Each metric is shortly described in the following sections. 

\subsubsection{Repeatability Score (RS)}
The repeatability score (RS) measures the quality of interest points and is the ratio between the number of points observed by both viewpoints and the total number of points \cite{Schmid2000-ri}. For a planar scene, the point correspondences between two camera views can be established by simply mapping points from one view to the other using a homography. To account for localization errors between two corresponding points, we define points to correspond if they are below a certain pixel distance defined as the correct distance $\rho$. To only evaluate points that are observable in both views, the repeatability measure will only include points in a region shared by the two viewpoints. Because the scale may change between two views, the repeatability rate depends on which camera view the points have been mapped to. The repeatability is therefore the average repeatability calculated in the view of each camera. 

\subsubsection{Localization Error (LE)}
The localization error (LE) is the average pixel distance between corresponding points. Only point-pairs with distances below $\rho$ are included in the calculation. Like repeatability rate, the localization error is the average error of corresponding points calculated in both camera views. 

\subsubsection{Homography estimation procedure}
\begin{figure*}
  \centering
    \includegraphics[width=0.95\textwidth]{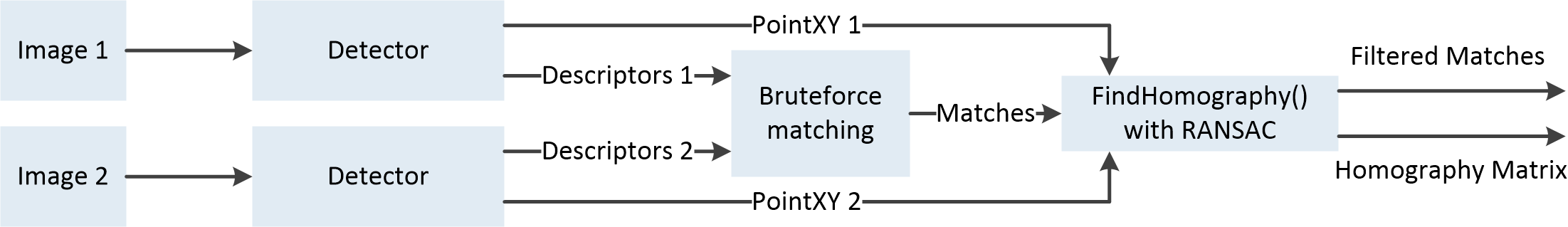}
  \caption{ Image matching for homography estimation. The detector generates point positions and descriptors for two images. Descriptors are matched using nearest neighbor (brute force) matching. The homography matrix is estimated using RANSAC based on matches and interest point positions.}
  \label{fig:homography_estimation_procedure}
\end{figure*}

The homography estimation procedure presented in Figure \ref{fig:homography_estimation_procedure} is commonly used in computer vision applications. The procedure may use any detector to select $N$ points from two images of the same (planar) scene. Descriptors from the two images are matched. In our procedure, we use nearest neighbor (brute force) matching with cross check. The homography is estimated with RANSAC \cite{Fischler1981-lv} using OpenCV. This uses matches and interest point positions to provide both a homography matrix and a filtered set of matches (that comply with the estimated homography). 

\subsubsection{Matching Score (MS)}
The matching score (MS) is the ratio between the correct matches and all points within the shared view.  A correct match is defined as two points that are nearest neighbors in descriptor space and separated by a pixel distance less than the correct distance $\rho$ after points have been transformed to the same view by the ground truth homography. 

\subsubsection{Homography Accuracy (HA)}
\begin{figure*}
  \centering
    \includegraphics[width=0.4\textwidth]{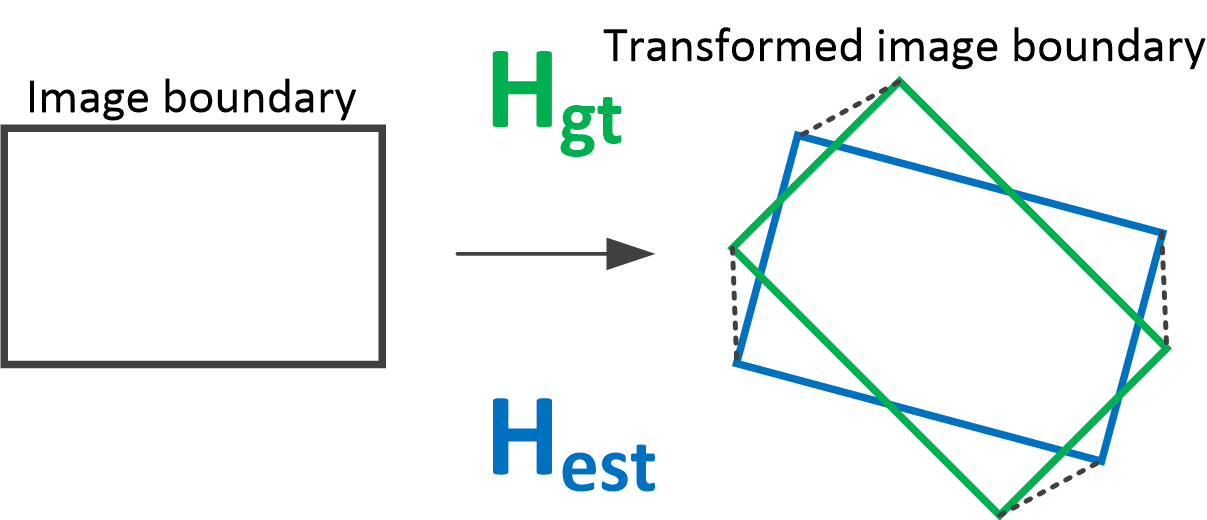}
  \caption{Homography error (HE) is the mean distance between corners of the target image after being transformed by 1) the ground truth homography $H_\text{gt}$ and 2) the estimated homography $H_\text{est}$. Black dashed lines represent the error. }
  \label{fig:metric_ha}
\end{figure*}

We define the homography error (HE) as the mean distance between target image corners transformed by the ground truth homography $H_\text{gt}$ and the estimated homography $H_\text{est}$. This is visualized in Figure \ref{fig:metric_ha} where image corners have been transformed by $H_\text{gt}$ and $H_\text{est}$. The distances between image corners are visualized by the dashed lines. 

The homography accuracy (HA) is the ratio between the number of correctly estimated homographies and the total number of homographies. To quantify if an estimated homography is correct, the HE should be less than a defined tolerance error $\epsilon$. Similar to SuperPoint, we measure HA at multiple tolerance values.

\subsection{Evaluations}
In this section, we compare UnsuperPoint to other detectors using the metrics specified in the previous section. The evaluation was performed on the full image sequence from the HPatch dataset \cite{Balntas2017-pn}. The dataset contains 57 illumination scenes and 59 viewpoint scenes. Each scene contains six images (one reference image and five target images) of a planar scene/surface and five transformations to map the reference frame to each of the five target frames. Each evaluated algorithm detected points for each frame and matched the reference frame to each target frame - creating a total of $57\cdot 5+59\cdot 5 = 580$ image pairs. Metrics were calculated and averaged for all image pairs. We defined two settings; $240\times320$ resolution images with $N = 300$ points and $480\times640$ resolution images with $N = 1000$ point. We have set the correct distance $\rho = 3$. To make a fair comparison of different detectors, we guaranteed that detectors always provided $N$ points in an image. To do this, we lowered the threshold of interest point detectors and then selected $N$ top points. 
Similar to \cite{DeTone2018-yw}, we used NMS for some evalations. If NMS was used, the top points were selected after NMS. 
For SIFT \cite{Lowe2004-qf}, SURF \cite{Bay2008-nf}, ORB \cite{Rublee2011-ns}, AKAZE \cite{Alcantarilla2013-au} and BRISK \cite{Leutenegger2011-ar} we used the implementation provided by OpenCV (v3.4.6-dev). 
We used the author-released Github versions for both SuperPoint \cite{SuperPointGit} and LF-Net \cite{LFNet}. For LF-net, we used both the indoor and outdoor models provided on the Github. 

\subsubsection{UnsuperPoint configurations}
In this section, we have evaluated different configurations of UnsuperPoint on $240\times320$ resolution images with $N = 300$. We present the benefit of 1) Interpolation, 2) uniform point predictions (UniformXY) and 3) decorrelation of descriptors (DecorrDesc). In Table \ref{tab:evaluate_configurations}, the metrics of the base model C0 and UnsuperPoint C4 are presented in the top and bottom row. Intermediate rows show the relative improvements in percentage to the base model C0. For localization, lower is better, and for other metrics, higher is better.

\begin{table*}[ht]
\centering
\begin{tabular}[b]{l|ccc|rrrrrr}
& \rotatebox[origin=l]{90}{\parbox[b]{2.0cm}{\  Interpolation}} 
& \rotatebox[origin=l]{90}{\parbox[b]{2.0cm}{\  UniformXY}} 
& \rotatebox[origin=l]{90}{\parbox[b]{2.0cm}{\  DecorrDesc}} 
& RS $\uparrow$ & LE $\downarrow$ &  \shortstack{HA $\uparrow$ \\ $\epsilon = 1$}  & \shortstack{HA $\uparrow$ \\ $\epsilon = 3$} & \shortstack{HA $\uparrow$ \\ $\epsilon = 5$} & MS $\uparrow$ \\
\hline
C0 base & & & & 0.633 & 0.898 & 0.519 & 0.812 & 0.871 & 0.458 \\
\hline
C1 & x & & & 1.1\% & 1.0\% & 0.0\% & 1.9\% & 2.0\% & 6.7\% \\
C2 & & x & & 1.9\% & 4.8\% & 15.3\% & 5.5\% & 4.0\% & 3.2\% \\
C3 & & & x & 0.2\% & 4.4\% & 8.3\% & 0.6\% & 2.6\% & 3.3\% \\
C4 UnsuperPoint & x & x & x & 1.9\% & 7.4\% & 7.3\% & 7.2\% & 5.7\% & 9.4\% \\
\hline
C4 UnsuperPoint & x & x & x & 0.612 & 0.991 & 0.521 & 0.831 & 0.902 & 0.452 \\
\end{tabular}
\caption{Relative improvements for configuration C1-C4 relative to base model C0. Top and bottom is the metrics for C0 base model C4 and SuperPoint. Interpolation improves especially matching score. UniformXY improves especially Homography estimation. Apart from repeatability, DecorrDesc improves overall performance. }
\label{tab:evaluate_configurations}
\end{table*}

The C1 model adds interpolation of descriptors to improve the matching score. The C2 model encourages a uniform distribution of point positions, which increases repeatability and homography estimation and reduces localization error. C3 adds decorrelation of descriptors to, apart from repeatability, improve overall performance. Finally, C4 uses all modules together. The final row presents the actual results of C4 which we denote as UnsuperPoint. 

\subsubsection{Point detector}
Table \ref{tab:point_detector} shows interest points for repeatability and localization error. Similar to \cite{DeTone2018-yw}, interest point metrics were calculated with NMS. 

\begin{table*}[ht]
\centering
\begin{tabular}{l|cc|cc}
& \multicolumn{2}{c|}{ Repeatability $\uparrow$ } &  \multicolumn{2}{c}{ Localization Error $\downarrow$} \\
& $240\times320$ & $480\times640$ & $240\times320$ & $480\times640$ \\
\hline
ORB & 0.532 & 0.525 & 1.429 & 1.430 \\
SURF & 0.491 & 0.468 & 1.150 & 1.244 \\
SURF\_EXT & 0.491 & 0.468 & 1.150 & 1.244 \\
AKAZ & 0.599 & 0.572 & 1.007 & 1.126 \\
BRISK & 0.566 & 0.505 & 1.077 & 1.207 \\
SIFT & 0.451 & 0.421 & 0.855 & 1.011 \\
SuperPoint & 0.631 & 0.593 & 1.109 & 1.212 \\
LF-Net (indoor) & 0.486 & 0.467 & 1.341 & 1.385 \\
LF-Net (outdoor) & 0.538 & 0.523 & 1.084 & 1.183 \\
UnsuperPoint & \textbf{\underline{0.645}} & \textbf{\underline{0.612}} &  \textbf{\underline{0.832}} &  \textbf{\underline{0.991}} \\
\end{tabular}
\caption{Repeatability (higher is better) and localization error (lower is better) for detectors with $240\times320$ and $480\times640$ resolution. 
\label{tab:point_detector}
}
\label{tab:evaluate_point_detector}
\end{table*}

\begin{figure*}
    \centering
    \begin{subfigure}[b]{0.48\textwidth}
        \includegraphics[width=\textwidth]{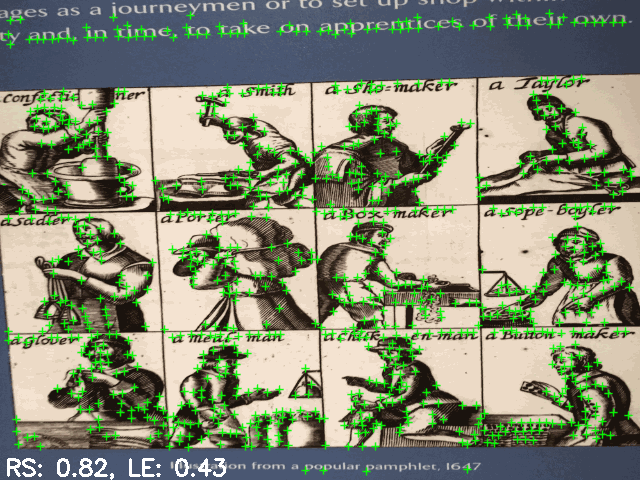}
    \end{subfigure}
    ~ 
    \begin{subfigure}[b]{0.48\textwidth}
        \includegraphics[width=\textwidth]{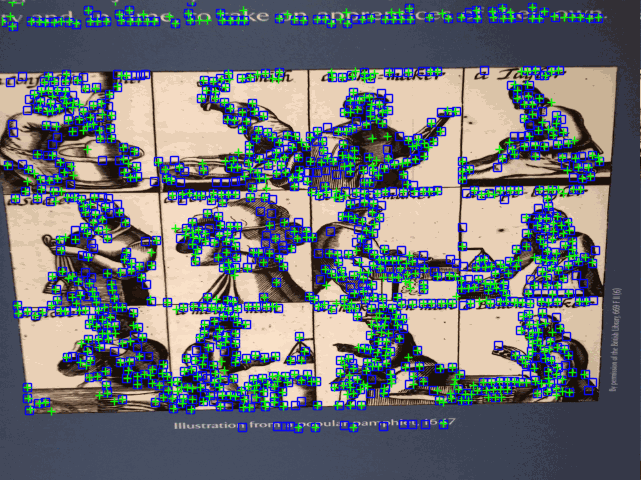}
    \end{subfigure}
    \caption{Interest point prediction on reference and target frame for small image motion. Predictions from the reference image are marked with green crosses (in both reference and target frame). Predictions from the target frame are marked with blue rectangles (only presented in target frame).}\label{fig:visualize_point_detections}
\end{figure*}

UnsuperPoint had both a high repeatability (similar to SuperPoint) and a low localization error (similar to SIFT). Figure \ref{fig:visualize_point_detections} visualizes interest points from two views of the same scene and demonstrates that the detector detects a combination of corners, blobs and edges. 

\subsubsection{Detector}
Table \ref{tab:evaluate_compare_detectors} presents the matching score and homography estimation with a tolerance threshold $\epsilon$ of 1, 3 and 5, respectively. In Table \ref{tab:evaluate_compare_detectors}, without NMS $\left(\text{NMS} = 0\right)$, UnsuperPoint has far more matches compared to other detectors. SuperPoint and UnsuperPoint were best for homography estimation with large tolerance errors. However, SIFT was better for estimating homographies with a low tolerance error $\epsilon = 1$. 

\begin{table*}
\centering
\begin{tabular}{l|ccc|c|ccc|c}
 & \multicolumn{4}{c|}{ $240\times320$, 300 points } & \multicolumn{4}{c}{ $480\times640$, 1000 points } \\
\hline
NMS = 0 & \multicolumn{3}{c|}{ HA $\uparrow$ } & & \multicolumn{3}{c|}{ HA $\uparrow$} & \\ 
& $\epsilon = 1$ & $\epsilon = 3$ & \multicolumn{1}{c|}{ $\epsilon = 5$} & MS $\uparrow$ & $\epsilon = 1$ & $\epsilon = 3$ & \multicolumn{1}{c|}{ $\epsilon = 5$} & MS $\uparrow$ \\
\hline
ORB & 0.131 & 0.422 & 0.540 & 0.218 & 0.286 & 0.607 & 0.710 & 0.204 \\
SURF & 0.397 & 0.702 & 0.762 & 0.255 & 0.421 & 0.745 & 0.812 & 0.230 \\
SURF\_EXT & 0.371 & 0.683 & 0.772 & 0.235 & 0.395 & 0.709 & 0.798 & 0.208 \\
AKAZE & 0.378 & 0.724 & 0.791 & 0.311 & 0.369 & 0.743 & 0.812 & 0.289 \\
BRISK & 0.414 & 0.767 & 0.826 & 0.258 & 0.300 & 0.653 & 0.746 & 0.211 \\
SIFT & \textbf{\underline{0.622}} & 0.845 & 0.878 & 0.304 & \textbf{\underline{0.602}} & 0.833 & 0.876 & 0.265 \\
SuperPoint & 0.491 & 0.833 & 0.893 & 0.318 & 0.509 & 0.834 & 0.900 & 0.281 \\
LF-net(ind.) & 0.183 & 0.628 & 0.779 & 0.326 & 0.231 & 0.679 & 0.803 & 0.287 \\
LF-net(outd.) & 0.347 & 0.728 & 0.831 & 0.296 & 0.400 & 0.745 & 0.834 & 0.241 \\
UnsuperPoint & 0.579 & \textbf{\underline{0.855}} & \textbf{\underline{0.903}} & \textbf{\underline{0.424}} & 0.493 & \textbf{\underline{0.843}} & \textbf{\underline{0.905}} & \textbf{\underline{0.383}} \\
\hline
NMS = 4 & & & & & & & &\\
\hline
SIFT & \textbf{\underline{0.636}} & 0.829 & 0.871 & 0.248 & \textbf{\underline{0.576}} & 0.807 & 0.855 & 0.213 \\
SuperPoint & 0.464 & 0.831 & 0.903 & 0.500 & 0.419 & 0.819 & \textbf{\underline{0.912}} & 0.441 \\
UnsuperPoint & 0.557 & \textbf{\underline{0.871}} & \textbf{\underline{0.921}} & \textbf{\underline{0.501}} & 0.521 & \textbf{\underline{0.831}} & 0.902 & \textbf{\underline{0.452}} \\

\end{tabular}
\caption{Homography estimation and matching score of detectors for low and medium resolution. 
}
\label{tab:evaluate_compare_detectors}
\end{table*}

SuperPoint was presented with an NMS module in the original paper. Bottom section of Table \ref{tab:evaluate_compare_detectors} presents also SIFT, SuperPoint and UnsuperPoint with NMS $\left(\text{NMS} = 4\right)$. Generally, SIFT does not improve with NMS. NMS improves SuperPoint and UnsuperPoint with far more matches and better homography estimation for large tolerance errors. 

\begin{figure*}
    \centering
    \begin{subfigure}[b]{0.98\textwidth}
        \includegraphics[width=\textwidth]{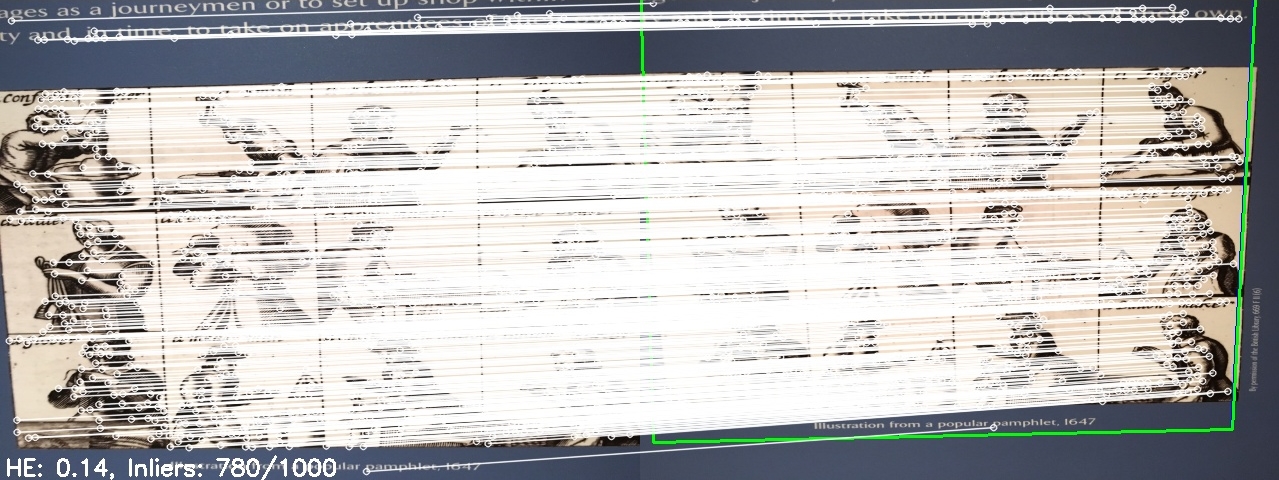}
    \end{subfigure}
      
    \centering
    \begin{subfigure}[b]{0.98\textwidth}
        \includegraphics[width=\textwidth]{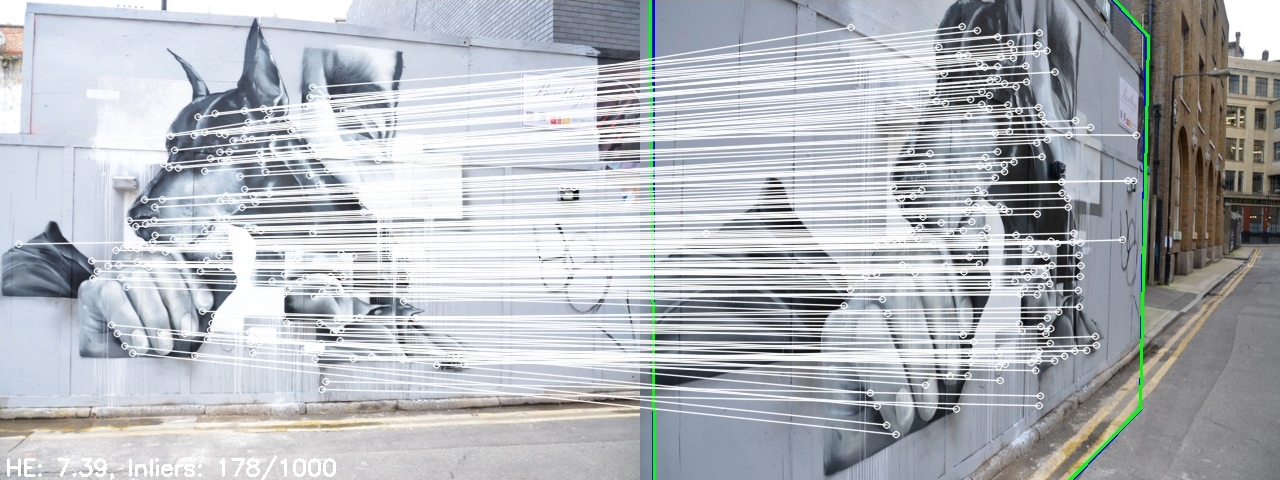}
    \end{subfigure}
    \caption{Filtered matches from UnsuperPoint for small and large motion examples. Matches are represented with lines. The ground truth homography (green) and estimated homography (blue) are visualized by transforming the reference image border into the target frame. }\label{fig:visualize_detector}
\end{figure*}

Figure \ref{fig:visualize_detector} presents filtered matches for UnsuperPoint to demonstrate the result of the detector after matching. Especially the bottom example demonstrates that UnsuperPoint was able to handle large perspective transformations caused by a change of viewpoint. 

\subsubsection{Speed}
Table \ref{tab:evaluate_speed} presents execution times of evaluated detectors on $240\times320$ and $480\times640$ resolution images. SuperPoint, UnsuperPoint and LF-Net were evaluated on GPU (GeForce Titan X), and remaining detectors were evaluated on CPU (Intel i7-7700HQ). 

SuperPoint is slightly faster than UnsuperPoint. However, the evaluation of SuperPoint does not include various post processing steps such as selecting points and interpolation. These steps have have not been optimized for speed and are therefore not included. 

\begin{table}[H]
\centering
\begin{tabular}{l|c|c}
Detectors & \shortstack{FPS @\\ $240\times320$} & \shortstack{FPS @ \\$480\times640$} \\
\hline
ORB & 91 & 33 \\
SIFT & 47 & 12 \\
SURF & 48 & 10 \\
SURF\_EXT & 48 & 10 \\
AKAZE & 100 & 18 \\
BRISK & 18 & 5 \\
SuperPoint $^\dagger$ & 167 & 67 \\
LF-NET $^\ddagger$ & 62 & 25 \\
UnsuperPoint, b1 & 119 & 65 \\
UnsuperPoint, b10 & 323 & 90 \\
\end{tabular}
\caption{Execution times of different detectors specified by frames per second (FPS) achieved on either CPU or GPU platforms. $^\dagger$ SuperPoint does not include interpolation and NMS. $^\ddagger$ LF-Net numbers from \cite{Ono2018-zq}.}
\label{tab:evaluate_speed}
\end{table}

%% file: sections/s07_discussion_conclusion_acknowledge.tex
\section{Discussion}
The used model is largely inspired by SuperPoint \cite{DeTone2018-yw} and is of similar performance in terms of speed, matching score and homography estimation for high tolerance errors. However, UnsuperPoint achieves better repeatability, lower localization error, better homography estimation for low tolerance errors and maintains matching score better without the use of NMS. Furthermore, we train the model from scratch and directly on MS COCO \cite{Lin2014-ee} images in a single training round, while SuperPoint requires synthetic data and four rounds of training. 

SIFT remains a good interest point detector with low localization error and the ability to estimate low homography errors $\epsilon = 1$ with a large margin. SuperPoint and UnsuperPoint are, however, better at estimating homographies when large errors are tolerated ($\epsilon = 3$ and $\epsilon = 5$), and they match far more points. Furthermore, SIFT runs only 12 fps on $480\times640$ images and is typically not considered for real-time applications. Moreover, it is patented and therefore not always suitable for commercial use. 

LF-Net presents a novel detector and training framework for self-supervised learning. The model is differentiable, learned from scratch in a single training round and fully rotational and scale invariant. Drawbacks of LF-net are that the detector and the descriptor do not share computations, and it requires an SfM-generated output to train. LF-net has shown state-of-the-art performance for general SfM applications, however, as demonstrated in the HPatch data, it is less powerful for image matching with small baseline image pairs. As also presented in the LF-Net paper, the performance of the detector drops when adding scale and rotational invariance. We argue that this has two causes. First, the extraction of image patches will restrict the visible area of the network to the patch. Secondly, an incorrect prediction of scale or rotation will damage the descriptor and cause the matching score to drop. For many applications (and especially in AR and VSLAM), the motion between frames in a video sequence is limited, and we can expect interest points to remain similar in scale and rotation. As demonstrated in this work and by SuperPoint, deep learning-based methods are powerful enough to learn some degree of invariance - without explicitly predicting scale and rotation. The gain for some applications is more matches and better homography estimation. 
In future work, we will test UnsuperPoint for VSLAM and/or AR. Furthermore, we use a simple backbone architecture and should therefore explore more advanced components such as residual connections \cite{He2016-wb}, dense convolutional layers \cite{Huang2016-vi}, squeeze-and-excitation \cite{Hu2017-or}, depth separable filters \cite{Chollet2016-ld, Howard2017-di} and skip connections from higher resolution features \cite{Long2015-nd}. 

\section{Conclusion}
We have established a framework for training a deep learning-based interest point detector using self-supervised learning. The framework and model use regression of point positions and a novel loss function to successfully train an interest point detector from scratch using no labels, no pseudo ground truth points and no Structure-from-Motion-generated outputs. Furthermore, we have investigated and successfully utilized a cost function to encourage a uniform distribution that may have utilization in other applications. The outcome is UnsuperPoint - a fast interest point detector with state-of-the-art performance.

\section{Acknowledgments}
This work is part of P. Christiansen’s Industrial PostDoc at EIVA A/S (eiva.com). The interest point detector is available under an EIVA license. This work is partly funded by the Innovation Fund Denmark (IFD) under File No. 8054-00002.

%% file: ms.bbl
\begin{thebibliography}{10}

\bibitem{LeCun2015-oh}
Yann LeCun, Yoshua Bengio, and Geoffrey Hinton.
\newblock Deep learning.
\newblock {\em Nature}, 521(7553):436--444, May 2015.

\bibitem{Krizhevsky2012-xg}
A~Krizhevsky, I~Sutskever, and Ge~Hinton.
\newblock ({AlexNet}) imagenet classification with deep convolutional neural
  networks.
\newblock {\em Adv. Neural Inf. Process. Syst.}, pages 1097--1105, 2012.

\bibitem{He2015-pm}
Kaiming He, Xiangyu Zhang, Shaoqing Ren, and Jian Sun.
\newblock Delving deep into rectifiers: Surpassing human-level performance on
  imagenet classification.
\newblock In {\em Proceedings of the {IEEE} international conference on
  computer vision}, pages 1026--1034, 2015.

\bibitem{Esteva2017-zy}
Andre Esteva, Brett Kuprel, Roberto~A Novoa, Justin Ko, Susan~M Swetter,
  Helen~M Blau, and Sebastian Thrun.
\newblock Dermatologist-level classification of skin cancer with deep neural
  networks.
\newblock {\em Nature}, 542(7639):115--118, February 2017.

\bibitem{Chung2016-zw}
Joon~Son Chung, Andrew Senior, Oriol Vinyals, and Andrew Zisserman.
\newblock Lip reading sentences in the wild.
\newblock November 2016.

\bibitem{Rajpurkar2017-ue}
Pranav Rajpurkar, Awni~Y Hannun, Masoumeh Haghpanahi, Codie Bourn, and Andrew~Y
  Ng.
\newblock {Cardiologist-Level} arrhythmia detection with convolutional neural
  networks.
\newblock July 2017.

\bibitem{Hartley2003-gk}
Richard Hartley and Andrew Zisserman.
\newblock {\em Multiple View Geometry in Computer Vision}.
\newblock Cambridge University Press, 2003.

\bibitem{Kendall2015-ql}
Alex Kendall, Matthew Grimes, and Roberto Cipolla.
\newblock Posenet: A convolutional network for real-time 6-dof camera
  relocalization.
\newblock In {\em Computer Vision ({ICCV)}, 2015 {IEEE} International
  Conference on}, pages 2938--2946, 2015.

\bibitem{Kendall2017-iv}
Alex Kendall and Roberto Cipolla.
\newblock Geometric loss functions for camera pose regression with deep
  learning.
\newblock In {\em Proceedings of the {IEEE} Conference on Computer Vision and
  Pattern Recognition}, pages 5974--5983, 2017.

\bibitem{DeTone2016-qe}
Daniel DeTone, Tomasz Malisiewicz, and Andrew Rabinovich.
\newblock Deep image homography estimation.
\newblock June 2016.

\bibitem{Luo2016-vm}
Wenjie Luo, Alexander~G Schwing, and Raquel Urtasun.
\newblock Efficient deep learning for stereo matching, 2016.

\bibitem{Wang2017-rh}
S~Wang, R~Clark, H~Wen, and N~Trigoni.
\newblock {DeepVO}: Towards end-to-end visual odometry with deep recurrent
  convolutional neural networks.
\newblock In {\em 2017 {IEEE} International Conference on Robotics and
  Automation ({ICRA})}, pages 2043--2050, May 2017.

\bibitem{Godard2017-hx}
Cl{\'e}ment Godard, Oisin Mac~Aodha, and Gabriel~J Brostow.
\newblock Unsupervised monocular depth estimation with left-right consistency.
\newblock In {\em Proceedings of the {IEEE} Conference on Computer Vision and
  Pattern Recognition}, pages 270--279, 2017.

\bibitem{Tareen2018-js}
S~A~K Tareen and Z~Saleem.
\newblock A comparative analysis of {SIFT}, {SURF}, {KAZE}, {AKAZE}, {ORB}, and
  {BRISK}.
\newblock In {\em 2018 International Conference on Computing, Mathematics and
  Engineering Technologies ({iCoMET})}, pages 1--10, March 2018.

\bibitem{Lowe2004-qf}
David~G Lowe.
\newblock Distinctive image features from {Scale-Invariant} keypoints, 2004.

\bibitem{Bay2008-nf}
Herbert Bay, Andreas Ess, Tinne Tuytelaars, and Luc Van~Gool.
\newblock {Speeded-Up} robust features ({SURF}).
\newblock {\em Comput. Vis. Image Underst.}, 110(3):346--359, June 2008.

\bibitem{Rublee2011-ns}
Ethan Rublee, Vincent Rabaud, Kurt Konolige, and Gary~R Bradski.
\newblock {ORB}: An efficient alternative to {SIFT} or {SURF}.
\newblock In {\em {ICCV}}, volume~11, page~2, 2011.

\bibitem{Alcantarilla2013-au}
Pablo Alcantarilla, Jesus Nuevo, and Adrien Bartoli.
\newblock Fast explicit diffusion for accelerated features in nonlinear scale
  spaces, 2013.

\bibitem{Leutenegger2011-ar}
Stefan Leutenegger, Margarita Chli, and Roland Siegwart.
\newblock {BRISK}: Binary robust invariant scalable keypoints.
\newblock In {\em 2011 {IEEE} international conference on computer vision
  ({ICCV})}, pages 2548--2555, 2011.

\bibitem{Triggs1999-lt}
Bill Triggs, Philip~F McLauchlan, Richard~I Hartley, and Andrew~W Fitzgibbon.
\newblock Bundle adjustment---a modern synthesis.
\newblock In {\em International workshop on vision algorithms}, pages 298--372,
  1999.

\bibitem{Mur-Artal2015-mm}
R~Mur-Artal, J~M~M Montiel, and J~D Tard{\'o}s.
\newblock {ORB-SLAM}: A versatile and accurate monocular {SLAM} system.
\newblock {\em IEEE Trans. Rob.}, 31(5):1147--1163, October 2015.

\bibitem{Mur-Artal2017-om}
R~Mur-Artal and J~D Tard{\'o}s.
\newblock {ORB-SLAM2}: An {Open-Source} {SLAM} system for monocular, stereo,
  and {RGB-D} cameras.
\newblock {\em IEEE Trans. Rob.}, 33(5):1255--1262, October 2017.

\bibitem{Mur-Artal2014-cd}
Ra{\'u}l Mur-Artal and Juan~D Tard{\'o}s.
\newblock Fast relocalisation and loop closing in keyframe-based {SLAM}.
\newblock In {\em 2014 {IEEE} International Conference on Robotics and
  Automation ({ICRA})}, pages 846--853, 2014.

\bibitem{Lhuillier2012-le}
Maxime Lhuillier.
\newblock Incremental fusion of {Structure-from-Motion} and {GPS} using
  constrained bundle adjustments.
\newblock {\em IEEE Trans. Pattern Anal. Mach. Intell.}, 34(12):2489--2495,
  December 2012.

\bibitem{Leutenegger2013-kj}
Stefan Leutenegger, Paul Furgale, Vincent Rabaud, Margarita Chli, Kurt
  Konolige, and Roland Siegwart.
\newblock Keyframe-based visual-inertial slam using nonlinear optimization.
\newblock {\em Proceedings of Robotis Science and Systems (RSS) 2013}, 2013.

\bibitem{Qin2018-wd}
T~Qin, P~Li, and S~Shen.
\newblock {VINS-Mono}: A robust and versatile monocular {Visual-Inertial} state
  estimator.
\newblock {\em IEEE Trans. Rob.}, 34(4):1004--1020, August 2018.

\bibitem{Han2015-vc}
X~Han, T~Leung, Y~Jia, R~Sukthankar, and {others}.
\newblock Matchnet: Unifying feature and metric learning for patch-based
  matching.
\newblock {\em Proc. IEEE}, 2015.

\bibitem{Brown2011-ml}
Matthew Brown, Gang Hua, and Simon Winder.
\newblock Discriminative learning of local image descriptors.
\newblock {\em IEEE Trans. Pattern Anal. Mach. Intell.}, 33(1):43--57, January
  2011.

\bibitem{Simo-Serra2015-lv}
Edgar Simo-Serra, Eduard Trulls, Luis Ferraz, Iasonas Kokkinos, Pascal Fua, and
  Francesc Moreno-Noguer.
\newblock Discriminative learning of deep convolutional feature point
  descriptors.
\newblock In {\em Proceedings of the {IEEE} International Conference on
  Computer Vision}, pages 118--126, 2015.

\bibitem{Zagoruyko2015-gz}
Sergey Zagoruyko and Nikos Komodakis.
\newblock Learning to compare image patches via convolutional neural networks.
\newblock In {\em Proceedings of the {IEEE} conference on computer vision and
  pattern recognition}, pages 4353--4361, 2015.

\bibitem{Balntas2016-nk}
Vassileios Balntas, Edward Johns, Lilian Tang, and Krystian Mikolajczyk.
\newblock {PN-Net}: Conjoined triple deep network for learning local image
  descriptors.
\newblock January 2016.

\bibitem{Tian2017-pw}
Yurun Tian, Bin Fan, and Fuchao Wu.
\newblock L2-net: Deep learning of discriminative patch descriptor in euclidean
  space.
\newblock In {\em Proceedings of the {IEEE} Conference on Computer Vision and
  Pattern Recognition}, pages 661--669, 2017.

\bibitem{Verdie2014-zw}
Yannick Verdie, Kwang~Moo Yi, Pascal Fua, and Vincent Lepetit.
\newblock {TILDE}: A temporally invariant learned {DEtector}.
\newblock November 2014.

\bibitem{Savinov2017-ew}
Nikolay Savinov, Akihito Seki, Lubor Ladicky, Torsten Sattler, and Marc
  Pollefeys.
\newblock Quad-networks: unsupervised learning to rank for interest point
  detection.
\newblock In {\em Proceedings of the {IEEE} conference on computer vision and
  pattern recognition}, pages 1822--1830, 2017.

\bibitem{Yi2016-on}
Kwang~Moo Yi, Eduard Trulls, Vincent Lepetit, and Pascal Fua.
\newblock {LIFT}: Learned invariant feature transform.
\newblock In {\em Computer Vision -- {ECCV} 2016}, Lecture Notes in Computer
  Science, pages 467--483. Springer, Cham, October 2016.

\bibitem{Jaderberg2015-xw}
Max Jaderberg, Karen Simonyan, Andrew Zisserman, and Koray Kavukcuoglu.
\newblock Spatial transformer networks.
\newblock June 2015.

\bibitem{Ono2018-zq}
Yuki Ono, Eduard Trulls, Pascal Fua, and Kwang~Moo Yi.
\newblock {LF-Net}: Learning local features from images.
\newblock In S~Bengio, H~Wallach, H~Larochelle, K~Grauman, N~Cesa-Bianchi, and
  R~Garnett, editors, {\em Advances in Neural Information Processing Systems
  31}, pages 6234--6244. Curran Associates, Inc., 2018.

\bibitem{DeTone2018-yw}
Daniel DeTone, Tomasz Malisiewicz, and Andrew Rabinovich.
\newblock Superpoint: Self-supervised interest point detection and description.
\newblock In {\em Proceedings of the {IEEE} Conference on Computer Vision and
  Pattern Recognition Workshops}, pages 224--236, 2018.

\bibitem{IoffeS14}
Sergey Ioffe and Christian Szegedy.
\newblock Batch normalization: Accelerating deep network training by reducing
  internal covariate shift.
\newblock {\em CoRR}, abs/1502.03167, 2015.

\bibitem{Bailo2018-cp}
Oleksandr Bailo, Francois Rameau, Kyungdon Joo, Jinsun Park, Oleksandr Bogdan,
  and In~So Kweon.
\newblock Efficient adaptive non-maximal suppression algorithms for homogeneous
  spatial keypoint distribution.
\newblock {\em Pattern Recognit. Lett.}, 106:53--60, April 2018.

\bibitem{Paszke2017-wo}
Adam Paszke, Sam Gross, Soumith Chintala, and Gregory Chanan.
\newblock Pytorch.
\newblock {\em Computer software. Vers. 0. 3}, 1, 2017.

\bibitem{Lin2014-ee}
Tsung-Yi Lin, Michael Maire, Serge Belongie, James Hays, Pietro Perona, Deva
  Ramanan, Piotr Doll{\'a}r, and C~Lawrence Zitnick.
\newblock Microsoft {COCO}: Common objects in context.
\newblock In {\em Computer Vision -- {ECCV} 2014}, pages 740--755. Springer
  International Publishing, 2014.

\bibitem{Kingma2014-gz}
Diederik~P Kingma and Jimmy Ba.
\newblock Adam: A method for stochastic optimization.
\newblock December 2014.

\bibitem{Schmid2000-ri}
Cordelia Schmid, Roger Mohr, and Christian Bauckhage.
\newblock Evaluation of interest point detectors.
\newblock {\em Int. J. Comput. Vis.}, 37(2):151--172, June 2000.

\bibitem{Fischler1981-lv}
Martin~A Fischler and Robert~C Bolles.
\newblock Random sample consensus: A paradigm for model fitting with
  applications to image analysis and automated cartography.
\newblock {\em Commun. ACM}, 24(6):381--395, June 1981.

\bibitem{Balntas2017-pn}
Vassileios Balntas, Karel Lenc, Andrea Vedaldi, and Krystian Mikolajczyk.
\newblock {HPatches}: A benchmark and evaluation of handcrafted and learned
  local descriptors.
\newblock In {\em Proceedings of the {IEEE} Conference on Computer Vision and
  Pattern Recognition}, pages 5173--5182, 2017.

\bibitem{SuperPointGit}
Daniel DeTone.
\newblock Superpoint.
\newblock
  \url{https://github.com/MagicLeapResearch/SuperPointPretrainedNetwork}.

\bibitem{LFNet}
Yuki Ono.
\newblock Lf-net.
\newblock \url{https://github.com/vcg-uvic/lf-net-release}.

\bibitem{He2016-wb}
Kaiming He, Xiangyu Zhang, Shaoqing Ren, and Jian Sun.
\newblock Identity mappings in deep residual networks.
\newblock In {\em Computer Vision -- {ECCV} 2016}, pages 630--645. Springer,
  Cham, October 2016.

\bibitem{Huang2016-vi}
Gao Huang, Zhuang Liu, Kilian~Q Weinberger, and Laurens van~der Maaten.
\newblock Densely connected convolutional networks.
\newblock August 2016.

\bibitem{Hu2017-or}
Jie Hu, Li~Shen, Samuel Albanie, Gang Sun, and Enhua Wu.
\newblock {Squeeze-and-Excitation} networks.
\newblock September 2017.

\bibitem{Chollet2016-ld}
Fran{\c c}ois Chollet.
\newblock Xception: Deep learning with depthwise separable convolutions.
\newblock October 2016.

\bibitem{Howard2017-di}
Andrew~G Howard, Menglong Zhu, Bo~Chen, Dmitry Kalenichenko, Weijun Wang,
  Tobias Weyand, Marco Andreetto, and Hartwig Adam.
\newblock {MobileNets}: Efficient convolutional neural networks for mobile
  vision applications.
\newblock April 2017.

\bibitem{Long2015-nd}
Jonathan Long, Evan Shelhamer, and Trevor Darrell.
\newblock Fully convolutional networks for semantic segmentation, 2015.

\end{thebibliography}
